\documentclass[conference]{IEEEtran}
\usepackage{cite}
\usepackage{graphicx}
\graphicspath{ {images/} }
\usepackage{amsmath,amssymb,amsfonts}
\usepackage{algorithmic}
\usepackage{textcomp}
\usepackage{url}
\usepackage{float}
\usepackage{tabularx}
\usepackage{upquote}
\usepackage{kantlipsum}
\usepackage{color}
\usepackage{hhline}
\usepackage{mathtools}

\begin{document}


\title{Illumination Invariant Foreground Object Segmentation using ForeGANs}


\author
{\IEEEauthorblockN{Maryam Sultana, Soon Ki Jung}
\IEEEauthorblockA{School of Computer Science and Engineering\\
Kyungpook National University\\
Daegu, South Korea\\
Email:maryam@vr.knu.ac.kr, skjung@knu.ac.kr}
}
\maketitle
\thispagestyle{plain}
\pagestyle{plain}





\begin{abstract}
The foreground segmentation algorithms suffer performance degradation in the presence of various challenges such as dynamic backgrounds, and various illumination conditions. To handle these challenges, we present a foreground segmentation method, based on generative adversarial network (GAN). We aim to segment foreground objects in the presence of two aforementioned major challenges in background scenes in real environments. To address this problem, our presented GAN model is trained on background image samples with dynamic changes, after that for testing the GAN model has to generate the same background sample as test sample with similar conditions via back-propagation technique. The generated background sample is then subtracted from the given test sample to segment foreground objects. The comparison of our proposed method with five state-of-the-art methods highlights the strength of our algorithm for foreground segmentation in the presence of challenging dynamic background scenario.
\end{abstract}

%
\IEEEpeerreviewmaketitle

\section{Introduction}
%
%
%
%
\IEEEPARstart{T}{he} fundamental steps in many computer vision and artificial intelligence applications involves background subtraction and foreground segmentation for the tasks of moving object detection. Foreground segmentation has further applications such as visual object tracking, video surveillance and salient motion detection. Background modeling is a crucial process, which describes the scene without the presence of any foreground objects. However, foreground segmentation is a process for extracting moving objects with prior knowledge of background scene.
Foreground segmentation is significantly affected by various challenges in background scene information, for instance, camera jitters, dynamic background, and sudden illumination variations. Despite that occlusion caused by foreground objects also, effects background model. Over the few decades, many techniques have been proposed in the literature to address problems of challenging background scenes for the tasks of foreground segmentation and evaluation \cite{bouwmans2017scene, bouwmans2018deep}. 

In this study, our primary focus is foreground segmentation in the presence of two major challenges in background scenes in real environments. 

\section{Related Work} \label{related work}
Background subtraction leads to foreground segmentation; it is a chicken-egg problem so many inclusive studies have been conducted to address this problem \cite{sultana2018unsupervised, javed2018spatiotemporal, javed2016motion}. A very famous and well-known method for background subtraction and foreground segmentation is \textit{Gaussian Mixture Model} (GMM) \cite{stauffer1999adaptive}. The basic idea of GMM is to use probability density functions based on mixture of Gaussians to model intensity variations in color at pixel level. Another very efficient and well-known technique for foreground segmentation along with background modeling is \textit{Robust Principal Component Analysis} (RPCA). Until now many techniques have been proposed based on RPCA method \cite{he2012incremental, DECOLOR, oreifej2013simultaneous, ye2015foreground, cao2016total, javed2017moving} for background subtraction and foreground segmentation. However, RPCA based techniques are mostly offline methods with high computational complexity and global optimization, which is a great challenge in these techniques. 
\section{Proposed Method} \label{Proposed} 
In this section, we describe each step of our proposed algorithm in detail, which we call "ForeGAN". Our proposed method is adopted from \cite{Sultana2018UnsupervisedRV} and we have enhanced this technique particularly for background identification by introducing scene-specific illumination information into DCGAN model \cite{radford2015unsupervised}. The proposed foreground segmentation technique has two phases. Phase 1.) Training of the ForeGAN model with background video sequences containing various illumination variations. Phase 2.) Testing of ForeGAN model with video sequences containing illumination variations including foreground objects.
\subsection{Training}\label{foregan}
A GAN model has two neural networks, a discriminator $D$ and a generator $G$. The objective of generator $G$ is to learn a distribution $p_{gen}$ over input data $X_t$ via mapping of $z$ samples through $G(z)$. This mapping facilitates the $1D$ vectors of input noise which is uniformly distributed and sampled from latent space $Z$ to the $2D$ image representation. In a GAN model discriminator, $D$ is a CNN model that maps a $2D$ image representation to a single value $D(\cdot)$. This single value $D(\cdot)$ of discriminator's output is considered as a probability that whether the input given to the discriminator $D$ was a fake image generated $G(z)$ by the generator $G$ or a real image $X$ sampled from training data $X_t$. The discriminator and the generator are simultaneously optimized via cross entropy loss functions in a following two-player minimax game with $\Gamma (D, G)$ as a value function:  
\begin{equation} \label{eq_fullgan}
\begin{split}
\min\limits_{G}\max\limits_{D}~\Gamma_{1}(G, D)= \mathbb{E}_{x\sim p_{data}(x)}[log(D(x))]\\
+ \mathbb{E}_{z\sim p_{z}(z)}[log(1 - D(G(z)))]. 
\end{split}
\end{equation}
The discriminator in GAN model is a decision maker entity which is trained to maximize the probability of assigning real training sample to actual input data and samples from $p_{gen}$ to the fake generated data. During the training process, the generator tries to improve itself by generating realistic images and the discriminator tries to identify the real and fake generated images.

\subsection{Testing}
During the phase:1 training, the generator learns the mapping from latent space representations $z$ to more realistic images. However inverse mapping in GAN is not a straightforward process; instead we need a different mechanism for this purpose. To achieve inverse mapping, a back-propagation method is applied to input data. It is the same back-propagation method which has also been used to understand and visualize neural network's learned features by inverting the network by updating gradients at input layer \cite{yeh2017semantic}. The loss functions to achieve back-propagation is discussed in detail in the next two sections.
Given a test image $x$ we aim to find that particular random noise $z$ in the latent space that was mapped to generate image $G(z)$ via back-propagation method. In order to find that specific $z$, we have to select an initial random sample $z_o$, from the latent space and reinforce it to the trained generator network to generate $G(z_o)$. 

\begin{equation}\label{FDL}
F(z_{\beta}) = \sum |x - G(z_{\beta})|.
\end{equation}
\section{Experiments} \label{experiments}

We presented results on two benchmark datasets Wallflower \cite{toyama1999wallflower} and I2R \cite{li2004statistical} for foreground segmentation. The testing of the proposed model is performed individually on all three datasets.  All the testing samples are resized to $64\times64$ and given as input to the all three models individually for validation. We set back-propagation steps to be $2000$ on all three datasets which are evaluated by using following $5$ metrics:
\begin{equation} \label{accu}
A = \frac{T_{p} + T_{n}}{T_{p} + F_{p} + F_{n} + T_{n}}.
\end{equation}
\begin{equation}
F =  \frac{2 (Pre\times Re)}{Pre + Re}.
\end{equation}
\begin{equation} \label{pre}
Pre =  \frac{T_{p}}{T_{p} + F_{p}}.
\end{equation}
\begin{equation} \label{re}
Re =  \frac{T_{p}}{T_{p} + F_{n}}.
\end{equation}
\begin{equation} \label{sp}
Sp =  \frac{T_{n}}{T_{n} + F_{p}},
\end{equation}
where $T_{p}$ is True positives, $T_{n}$ is True negatives, $F_{p}$ is False positives, $F_{n}$ is False negatives, $A$ is Accuracy, $F$ is F-Measure score, $Pre$ is Precision, $Re$ is Recall and $Sp$ is Specificity. For better foreground segmentation the aim of the metrics (defined in equations \eqref{accu}-\eqref{sp}) is to achieve maximum values in all of $5$ metrics.
\begin{figure}
	\centering
	\includegraphics[scale=0.55]{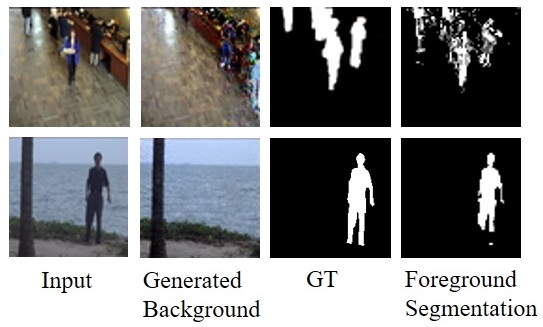}
	\caption{Performance comparison of ForeGAN method on all tow datasets based on background images generated by proposed GAN model along with foregrounds segmentation.row: (1) video sequence "Bootstrap" both from Wallflower dataset, row: (2) video sequence "WaterSurface" both from I2R dataset, }
	\label{fig_ICDD}
\end{figure}
we have presented the comparison of ForeGAN model with $5$ state-of-the-art methods in context to foreground segmentation. By using original implementations of the authors, we have compared our proposed method with GRASTA \cite{he2012incremental}, DECOLOR \cite{DECOLOR}, 3TD \cite{oreifej2013simultaneous}, RAMAR \cite{ye2015foreground} and TVRPCA \cite{cao2016total} and evaluated the results of our proposed method on two benchmark datasets. Wallflower dataset and I2R dataset has challenges like dynamic background changes, illuminations conditions, and camouflage objects. It can be seen in Table \ref{f_2datasets} that our ForeGAN model on average has achieved the highest F-measure score in both datasets.
\begin{table}
\centering
\caption{Comparison of proposed ForeGAN methods by using F-measure score on three datasets. The first highest and the second highest scores for each dataset is shown in red and blue color respectively.}
	\scalebox{0.66}{
		\label{f_2datasets}
		\begin{tabular}{l*{6}{c}r}
			\hline
			Datasets  & GRASTA \cite{he2012incremental}  & DECOLOR \cite{DECOLOR} & 3TD \cite{oreifej2013simultaneous} & RAMAR \cite{ye2015foreground} & TVRPCA \cite{cao2016total} &ForeGAN \\
			\hline
			Wallflower & 0.3303  & 0.5904 & 0.7559 &  \textcolor{blue}{0.8004} & 0.6170  & \textcolor{red}{0.8225}\\
			I2R      & 0.5489  & 0.7401 & 0.7251 & \textcolor{blue}{0.7505} & 0.6954  & \textcolor{red}{0.7782} \\
			\hline
		\end{tabular}
		}
\end{table}
\section{Conclusion} \label{con}
In this study, we present the foreground segmentation algorithm based on Generative Adversarial Network (GAN). Our goal is to segment foreground objects in the presence of two major challenges in background scenes in real environments. The two challenges are dynamic background and camouflage conditions. 
For this problem, we have presented a solution based on GAN working on the principle of generating background image samples with specific conditions. 
The comparison of our proposed method with five state-of-the-art methods highlights the strength of our algorithm for foreground segmentation in the presence of challenging illumination conditions and dynamic background scenario.
\section{acknowledgements}
This research was supported by Development project of leading technology for future vehicle of the business of Daegu metropolitan city (No. 20171105).
\bibliographystyle{IEEEtran}
\bibliography{mybifile}
\end{document}